\documentclass{article}
\usepackage{spconf,amsmath,graphicx}
\usepackage{tikz,pgfplots}
\usetikzlibrary{plotmarks}
\usepackage{adjustbox}
\usepackage{amssymb}
\usepackage{amsmath}
\usepackage{todonotes}
\usepackage{hyperref}
\usepackage[caption=false,font=footnotesize]{subfig}
\usetikzlibrary{patterns}
\usepackage{makecell}
\newcolumntype{x}[1]{>{\centering\arraybackslash}p{#1}}

\definecolor{darkgreen}{RGB}{47,109,79}
\definecolor{rosso}{RGB}{220,57,18}
\DeclareMathOperator*{\argmin}{arg\,min}
\title{Distributed One-class Learning}
\name{Ali Shahin Shamsabadi$^{\star}$, Hamed Haddadi$^{\dagger}$, Andrea Cavallaro$^{\star}$}
\address{$^{\star}$Queen Mary University of London,$^{\dagger}$Imperial College London}
\begin{document}
\ninept
\maketitle
\begin{abstract}
We propose a cloud-based filter trained to block third parties from uploading privacy-sensitive images of others to online social media. The proposed filter uses {Distributed One-Class Learning}, which decomposes the cloud-based filter into multiple one-class classifiers. Each one-class classifier captures the properties of a class of privacy-sensitive images with an autoencoder. The multi-class filter is then reconstructed by combining the parameters of the one-class autoencoders. The training takes place on edge devices (e.g.~smartphones) and therefore users do not need to upload their private and/or sensitive images to the cloud. A major advantage of the proposed filter over existing distributed learning approaches is that users cannot access, even indirectly, the parameters of other users. Moreover, the filter can cope with the imbalanced and complex distribution of the image content and the independent probability of addition of new users. We evaluate the performance of the proposed distributed filter using the exemplar task of blocking a user from sharing privacy-sensitive images of other users. In particular, we validate the behavior of the proposed multi-class filter with non-privacy-sensitive images, the accuracy when the number of classes increases, and the robustness to attacks when an adversary user has access to privacy-sensitive images of other users. 
\end{abstract}
%\vspace{-0.2cm}
%%%%%%%%%%%%%%%%%%%%%%%%%%%%%%%
\section{Introduction}
\label{sec:intro}

Unauthorized sharing of potentially privacy-sensitive images of other users is an increasingly important privacy challenge in online social media. To protect privacy, a cloud sharing images should filter uploaded content to prevent the sharing of unauthorized (or undesired) privacy-sensitive images. However, to reach this goal a service provider would need to access the privacy-sensitive images themselves in order to produce a hash or to train a filter. This centralized learning solution only shifts and does not solve the problem of maintaining certain images private~\cite{facebook}. 
 
This privacy challenge could be addressed by distributed learning~\cite{shokri2015privacy,mcmahan2016communication}, where each user (i.e.~each edge device) uploads to the service provider only the \emph{parameters} of their machine learning model (from a {\em local} copy of the service provider's learning model), not their raw images. The service provider then fine-tunes the {\em global} learning models by combining the values of the parameters of each user. Finally, the updated parameters are downloaded by the device of each user, which would consequently update their local learning models and upload the updated parameters back to the cloud. This iterative process stops when a certain classification accuracy is achieved. 

While distributed learning can be considered more privacy-friendly than centralized learning (as a result of uploading parameters of local models instead of raw image dataset), it has important limitations. First, the parameters of each user are not only shared with the service provider, but also (indirectly) with the other users when the shared parameters are downloaded from the cloud. Through the analysis of the parameters during the training phase an adversary user can recover information about the training data of other users~\cite{hitaj2017deep,zhang2016understanding}. Second, the user should have access to data of several classes as each user updates the parameters of a local copy of a multi-class classifier. This is in contrast to our scenario where each user only has access to data of their own class(es). Third, a new user cannot join the system during the test phase when other users are using the service, and therefore scalability is an issue.

To tackle these limitations, we propose \emph{DOCL}, a Distributed One-Class Learning approach that decomposes the global filter down to $N$ one-class classifiers distributed among the users. Each user trains an autoencoder as a classifier on their privacy-sensitive data. The autoencoder learns to reconstruct its privacy-sensitive training image with minimum error. The global filter then aggregates all the one-class autoencoders and, to discriminate between classes and block privacy-sensitive images, measures the dissimilarity between new images uploaded by (other) users and the reconstructions of the autoencoders. 

The proposed filter has several desirable properties. The training of each one-class classifier is {\em independent} from that of the other classifiers and therefore training data and parameters are not shared among users, thus preserving the privacy of training data against adversary users. Moreover, as the global filter is decomposed into a series of simple one-class classifiers trained by the users themselves, even the service provider has no access to the privacy-sensitive images of the users. In addition to the above, the proposed filter can cope with {\em imbalanced training data}, which is an important property as the number of training images for the different classes is likely to be different. In particular, when the imbalance of the training data increases, the performance of {one-class classifiers} is almost stable, whereas that of, for example, multi-class classifiers (even binary classifiers) decreases~\cite{bellinger2012one}. Finally, the size of the global filter depends on the number of users, which can be easily increased by uploading their one-class classifiers trained on their privacy-sensitive images to the cloud at any point. 

%%%%%%%%%%%%%%%%%%%%%%%%%%%%%%%%
\section{The distributed learning model}
\label{sec:proposed}

\subsection{The local classifier: training at the edge}

One-class classifiers or data descriptors~\cite{tax2001one} learn to distinguish the target class from outlier classes (i.e.~when only data of the target class are available) using density estimation~\cite{duda1973pattern}, data reconstruction~\cite{manevitz2007one} or closed boundary estimation~\cite{tax2004support,tax2003kernel,scholkopf2001estimating}. In our proposed framework, DOCL, each user trains as classifier a one-class autoencoder~\cite{manevitz2007one}, a three-layer parametric neural network with an input layer, a data representation layer, and an output layer~\cite{hinton2006reducing, vincent2008extracting, shamsabadi2017new}. 

%  \begin{figure}[t!]
%   \includegraphics[width=\columnwidth]{imgs/Cloud.pdf}
%     \caption{Service provider feed the new image ($I^t$) to FE to extract feature ($x^t$). $x^t$ is then fed to the $N$ one-class autoencoders. The dissimilarity ($d$) between $x^t$ and its reconstructions $\hat{x}^t$ are computed. Service provider then checks the minimum dissimilarity with its corresponding APA for assigning the class of new uploaded image.   } 
%   \label{fig:cloud}
%  \end{figure}

For simplicity but without loss of generality, let the number of classes correspond to the number of users, $N$. 
Let the users $\{u_0,u_1,...,u_{N-1}\}$ train independently $N$ one-class autoencoders on their privacy-sensitive image datasets $\{I_0,I_1,...,I_{N-1}\}$. Each $u_i$ feeds their set $I_i=\{\mathbf{I}_{i,0},\mathbf{I}_{i,1},...,\mathbf{I}_{i,j},...,\mathbf{I}_{i,K_i-1}\}$ to a  ResNet~\cite{he2016deep}, which is already pre-trained by the service provider in the cloud. Note that the cardinality of each set $|I_i|=K_i$ may considerably differ across users. For each $I_i$, ResNet generates the corresponding feature set, $X_i$, defined as:    
\begin{equation}
X_i=\{\mathbf{x}_{i,0},\mathbf{x}_{i,1},...,\mathbf{x}_{i,j},...,\mathbf{x}_{i,K_i-1}\},
\end{equation}
where $\mathbf{x}_{i,j} \in \mathbb{R}^{D} $ and $D$ is the feature dimension.  

The parametric {\em encoder} of user $u_i$ obtains  $\mathbf{h}_{i,j} \in \mathbb{R}^{M}$, a feature representation whose dimension $M<D$, by applying a Rectifier Linear Unit $f(\cdot)$~\cite{glorot2011deep} on the linear combination of the elements of feature $\mathbf{x}_{i,j}$:
\begin{equation}\label{eq:enc}
\mathbf{h}_{i,j} = f(\mathbf{W}_{i}\mathbf{x}_{i,j}+\mathbf{b}_{i})\quad \forall j=0,1,...,K_i-1,
\end{equation}
where $\mathbf{W}_{i}\in \mathbb{R}^{M \times D}$ and $\mathbf{b}_{i} \in \mathbb{R}^M$ are the parameters of the encoder for user $u_i$. Then the parametric {\em decoder} of user $u_i$ maps feature representation $\mathbf{h}_{i,j}$ to the feature reconstruction $\mathbf{\hat{x}}_{i,j} \in \mathbb{R}^{D}$ by applying a sigmoid function $g(\cdot)$~\cite{han1995influence} on the linear combination of the elements of $\mathbf{h}_{i,j}$:
\begin{equation}\label{eq:dec}
\mathbf{\hat{x}}_{i,j} = g(\mathbf{W}^{'}_{i}\mathbf{h}_{i,j}+\mathbf{b}^{'}_{i})\quad \forall j=0,1,...,K_i-1,
\end{equation}
where $\mathbf{W}^{'}_{i} \in \mathbb{R}^{D \times M}$ and $\mathbf{b}^{'}_{i} \in \mathbb{R}^ D$ are the decoder's parameters for $u_i$.

Let $C(\cdot)$ represent the binary cross-entropy differences among features and feature reconstructions:
\begin{equation}
C(\mathbf{x}_{i,j},\mathbf{\hat{x}}_{i,j})= -\sum_{l=1}^{D}{x_{i,j}}(l)\log{{\hat{x}_{i,j}}(l)}+(1- x_{i,j})\log(1- {\hat{x}_{i,j}}(l)),
\end{equation}
where ${x_{i,j}}(l)$ and ${\hat{x}_{i,j}}(l)$ are the $l$-th elements of $\mathbf{x}_{i,j}$ and $\mathbf{\hat{x}}_{i,j}$, respectively.
The final parameters for $u_i$, $\{\mathbf{W}_{i}^{*},\mathbf{W}^{'*}_i, \mathbf{b}_i^{*}, \mathbf{b}^{'*}_i\}$, are learned with the following optimization:
\begin{equation}
\label{eq:loss}
\{\mathbf{W}_{i}^*,\mathbf{W}^{'*}_i, \mathbf{b}_i^*, \mathbf{b}^{'*}_i\} = \argmin_{\{\mathbf{W}_{i},\mathbf{W}^{'}_{i}, \mathbf{b}_{i}, \mathbf{b}^{'}_{i}\}} \sum_{j=0}^{K_i-1}C(\mathbf{x}_{i,j},\mathbf{\hat{x}}_{i,j})
\end{equation}
and are obtained using ADADELTA~\cite{zeiler2012adaptive}. These parameters are then uploaded to the cloud for the global filter to populate the $N$-class classifier composed of $N$ one-class\footnote{Note that this is the worst case of imbalanced training data, where data of each user belongs to one class only.} autoencoders. Note that unlike~\cite{shokri2015privacy,mcmahan2016communication}, in our case the parameters of each user are {\em not} downloaded by other users.

Moreover, each user $u_i$ uploads to the global filter also the parameters that describe the distribution of the differences, $d(\mathbf{x}_{i,j},\mathbf{\hat{x}}_{i,j})$,  between the input features and the reconstructed features for all training images, using the final parameters of their autoencoder. These differences are computed as:
\begin{equation}
\label{eq:dis}
\centering		d(\mathbf{x}_{i,j},\mathbf{\hat{x}}_{i,j}) = \sum_{l=1}^{D}||{x}_{i,j}(l)-{\hat{x}_{i,j}(l)}||_{2}^{2},\quad\forall j=0,1,..,K_i-1,
\end{equation}
where ${x_{i,j}}(l)$ and ${\hat{x}_{i,j}(l)}$ are, respectively, the $l$-th elements of feature vector of a training image and that of its reconstruction. The edge device of each user, $u_i$, fits the distribution $d_{i,j}=d(\mathbf{x}_{i,j},\mathbf{\hat{x}}_{i,j})$ $\forall j=0,1,..,K_i-1$ with a normal distribution and uploads its mean, $\mu_i$, and standard deviation, $\sigma_i$, to the global filter in the cloud.

%%%%%%%%%%%%%%%%%%%%%%%%%%%%%%%
\subsection{The global filter}

 \begin{figure}[t]
   \includegraphics[width=\columnwidth]{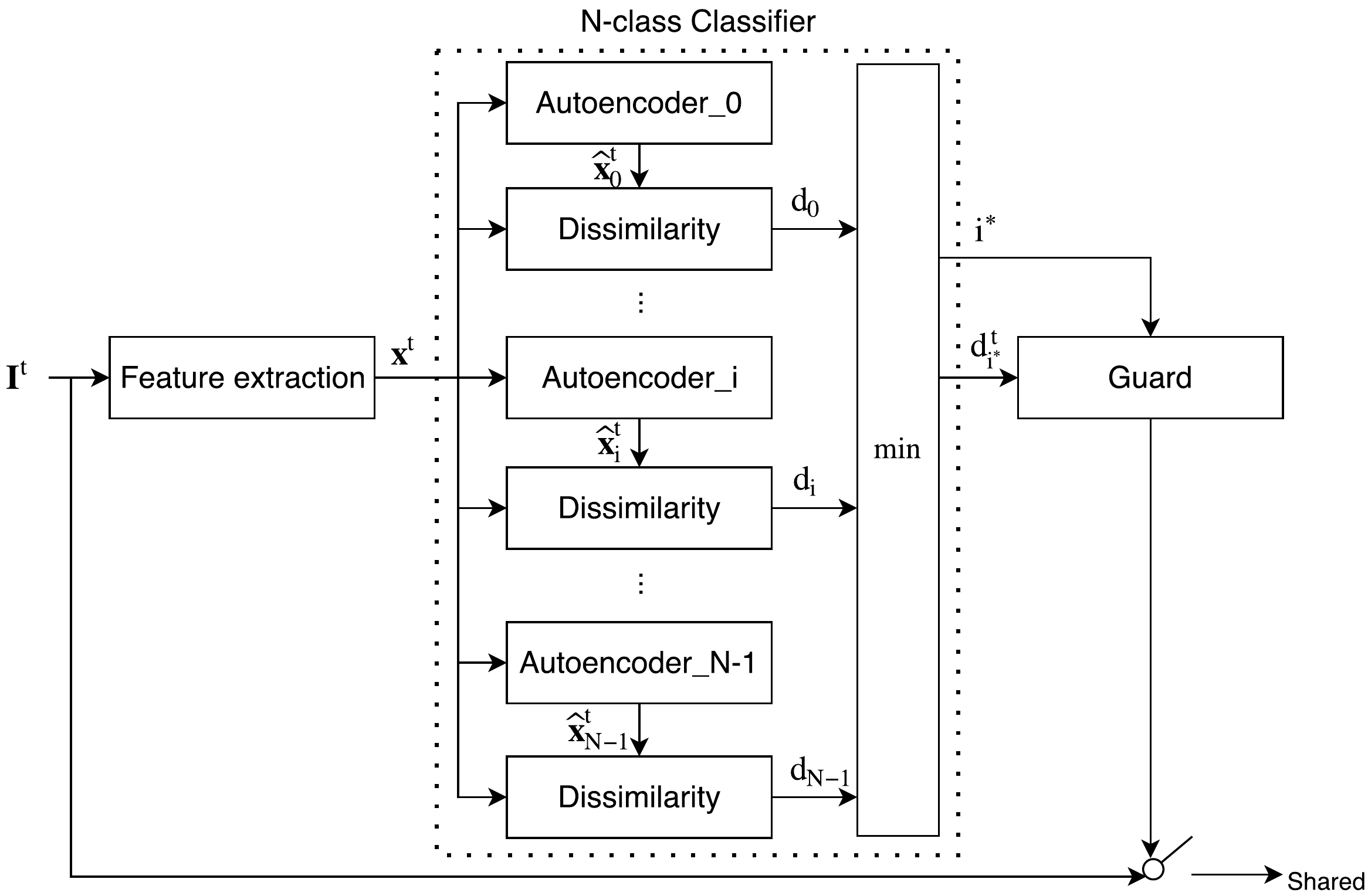}
     \caption{Block diagram of the global filter. The service provider extract the features $\mathbf{x}^t$ from a newly uploaded image $\mathbf{I}^t$ passes them through the $N$ one-class autoencoders, each trained independently by users. The minimum dissimilarity between $\mathbf{x}^t$ and its reconstructions $\mathbf{\hat{x}}^{t}$ is then analyzed by the Guard module prior to sharing $\mathbf{I}^t$.}
 % \vspace{-10px}
 \label{fig:filter}
 \end{figure}

%%%%%%%%%%%%%%%%%%%%%%%% Start table of datasets
\begin{table*}[t!]
\centering
\resizebox{\textwidth}{!}{ 
\begin{tabular}{|c|c|c|c|c|c|c|c|c|c|c|}

 \hline
 & $u_0$ & $u_1$ & $u_2$ & $u_3$ & $u_4$ & $u_5$ & $u_6$ & $u_7$ & $u_8$ & $u_9$ \\
 \hline
 \hline
IMDB&\adjustimage{height=1cm,valign=m}{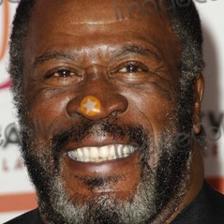} 402&\adjustimage{height=1cm,valign=m}{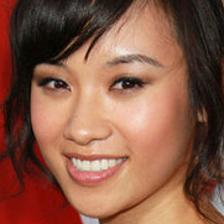}414&\adjustimage{height=1cm,valign=m}{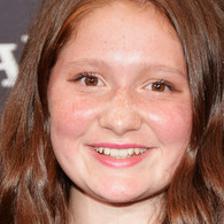} 347&\adjustimage{height=1cm,valign=m}{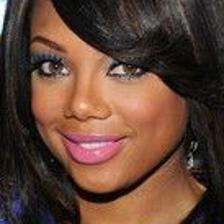}528 & \adjustimage{height=1cm,valign=m}{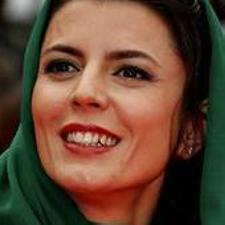}559&\adjustimage{height=1cm,valign=m}{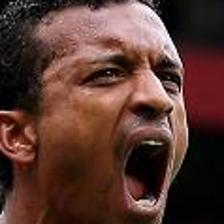} 198& \adjustimage{height=1cm,valign=m}{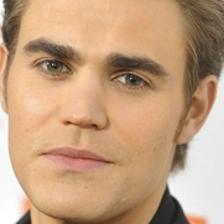} 414& \adjustimage{height=1cm,valign=m}{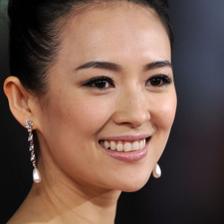}704& \adjustimage{height=1cm,valign=m}{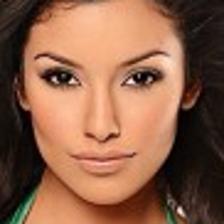} 619& \adjustimage{height=1cm,valign=m}{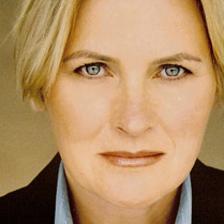}554\\
\hline
CIFAR-10&\adjustimage{height=1cm,valign=m}{airplane}5000 &\adjustimage{height=1cm,valign=m}{car}5000&\adjustimage{height=1cm,valign=m}{bird}5000&\adjustimage{height=1cm,valign=m}{cat} 5000& \adjustimage{height=1cm,valign=m}{dear}5000&\adjustimage{height=1cm,valign=m}{dog} 5000& \adjustimage{height=1cm,valign=m}{frog}5000 & \adjustimage{height=1cm,valign=m}{horse}5000& \adjustimage{height=1cm,valign=m}{ship} 5000& \adjustimage{height=1cm,valign=m}{truck}5000\\
\hline
MNIST&\adjustimage{height=1cm,valign=m}{zero}5923
&\adjustimage{height=1cm,valign=m}{one}6742&\adjustimage{height=1cm,valign=m}{two}5958&\adjustimage{height=1cm,valign=m}{three} 6131& \adjustimage{height=1cm,valign=m}{four}5842&\adjustimage{height=1cm,valign=m}{five} 5421& \adjustimage{height=1cm,valign=m}{six}5918 & \adjustimage{height=1cm,valign=m}{seven}6265& \adjustimage{height=1cm,valign=m}{eight} 5851& \adjustimage{height=1cm,valign=m}{nine}5949\\
\hline

\end{tabular}
}
\caption{A sample image for each class and the number of training images from the IMDB, CIFAR-10, and MNIST datasets.}
\label{tab:mnist}
\end{table*}

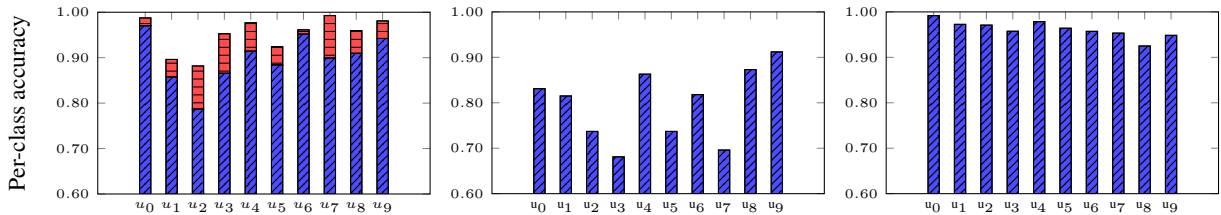
\begin{figure*}[t!]
\centering
\subfloat{
\begin{tikzpicture}
\begin{axis}[                        
tiny,
width=6cm,
height = 4cm,
bar width=0.15cm,
ybar stacked,
ylabel={Per-class accuracy},
label style={font=\tiny},
ticklabel style = {font=\tiny},
enlarge x limits=0.2,
legend style={at={(0.5,1.3)},anchor=north,legend columns=-1, transpose legend},
legend image code/.code={%
\draw[#1] (0cm,-0.1cm) rectangle (0.2cm,0.2cm);
},
y label style={font=\small, at={(axis description cs:.1,.5)},anchor=south},
y tick label style={
/pgf/number format/.cd,
fixed,
fixed zerofill,
precision=2,
/tikz/.cd
},
symbolic x coords={$u_0$, $u_1$, $u_2$, $u_3$, $u_4$, $u_5$, $u_6$, $u_7$, $u_8$, $u_9$},
xtick=data, ymin=0.6, ymax=1,
%		nodes near coords,
every node near coord/.append style={font=\tiny},
nodes near coords align={vertical},
]

\addplot[fill=blue!70!white,postaction={pattern=north east lines}] coordinates {($u_0$,0.9701) ($u_1$,0.8574) ($u_2$,0.7867)($u_3$,0.8655) ($u_4$,0.9141)($u_5$,0.8838)($u_6$,0.9516)($u_7$,0.8991)($u_8$,0.9095)($u_9$,0.9422)};\label{s}
\addplot[fill=red!70!white,postaction={pattern=horizontal lines}] coordinates {($u_0$,0.0174) ($u_1$,0.0387) ($u_2$,0.0951)($u_3$,0.0871) ($u_4$,0.0626)($u_5$,0.0404)($u_6$,0.0097)($u_7$,0.0937)($u_8$,0.0501)($u_9$,0.0387)};\label{b}
;
\end{axis}
\end{tikzpicture}
}
\subfloat{
\begin{tikzpicture}
\begin{axis}[
tiny,
width=6cm,
height=4cm,
bar width=0.15cm,
ybar stacked,
label style={font=\tiny},
ticklabel style = {font=\tiny},
enlarge x limits=0.2,
legend style={at={(0.5,1.3)},anchor=north,legend columns=-1, draw=none},
legend image code/.code={%
\draw[#1] (0cm,-0.1cm) rectangle (0.2cm,0.2cm);
},
y label style={font=\small, at={(axis description cs:.1,.5)},anchor=south},
y tick label style={
/pgf/number format/.cd,
fixed,
fixed zerofill,
precision=2,
/tikz/.cd
},
symbolic x coords={u$_0$, u$_1$, u$_2$, u$_3$, u$_4$, u$_5$, u$_6$, u$_7$, u$_8$, u$_9$},
xtick=data, ymin=0.6, ymax=1,
%		nodes near coords,
every node near coord/.append style={font=\tiny},
nodes near coords align={vertical},
]
\addplot[fill=blue!70!white,postaction={pattern=north east lines}] coordinates {(u$_0$,0.8310) (u$_1$,0.815) (u$_2$,0.737)(u$_3$,0.681) (u$_4$,0.863)(u$_5$,0.737)(u$_6$,0.818)(u$_7$,0.696)(u$_8$,0.873)(u$_9$,0.912)};
\end{axis}
\end{tikzpicture}
}
\subfloat{
\begin{tikzpicture}
\begin{axis}[
tiny,
width=6cm,
height=4cm,
bar width=0.15cm,
ybar stacked,
label style={font=\tiny},
ticklabel style = {font=\tiny},
enlarge x limits=0.2,
legend style={at={(0.5,1.3)},anchor=north,legend columns=-1, draw=none},
legend image code/.code={%
\draw[#1] (0cm,-0.1cm) rectangle (0.2cm,0.2cm);
},
y label style={font=\small, at={(axis description cs:.1,.5)},anchor=south},
y tick label style={
/pgf/number format/.cd,
fixed,
fixed zerofill,
precision=2,
/tikz/.cd
},
symbolic x coords={u$_0$, u$_1$, u$_2$, u$_3$, u$_4$, u$_5$, u$_6$, u$_7$, u$_8$, u$_9$},
xtick=data, ymin=0.6, ymax=1,
%		nodes near coords,
every node near coord/.append style={font=\tiny},
nodes near coords align={vertical},
]
\addplot[fill=blue!70!white,postaction={pattern=north east lines}] coordinates {(u$_0$,0.9918) (u$_1$,0.9726) (u$_2$,0.9709)(u$_3$,0.9574) (u$_4$,0.9786)(u$_5$,0.9641)(u$_6$,0.9572)(u$_7$,0.9533)(u$_8$,0.9250)(u$_9$,0.9484)};                 
\end{axis}
\end{tikzpicture}
}
%\vspace{-10px}
\caption{Per-class accuracy of the global filter on IMDB (left), CIFAR-10 (middle) and MNIST (right) images. Note that for  IMDB we used two autoencoders with representation layers of size 32 \ref{s} and 132 \ref{b}.}
%\vspace{-10px}
\label{fig:performance-cifar}
\end{figure*}

After aggregating the $N$ locally trained one-class autoencoders in the global filter, users can use the service by uploading their images (see Fig.~\ref{fig:filter}). 

Let a user upload image $\mathbf{I}^{t}$ during the test phase. Prior to sharing that image, the service provider checks its legitimacy  by feeding the features $\mathbf{x}^{t}$ generated from $\mathbf{I}^{t}$ to the global filter, which generates the {reconstructed feature vector} $\mathbf{\hat{x}}^{t}_{i}$ for each autoencoder:
 \begin{equation}
   \mathbf{\hat{x}}^{t}_{i} = g\bigg(\mathbf{W}^{'*}_{i}\big(f(\mathbf{W}^{*}_{i}{\mathbf{x}^{t}}+\mathbf{b}^{*}_{i})\big)+\mathbf{b}^{'*}_{i}\bigg) \quad \forall i=1,2,..,N.  
 \end{equation}

Each autoencoder reconstructs differently images of the same class as its training set and images of another class~\cite{manevitz2007one}. Hence the service provider quantifies the dissimilarity between the features of the uploaded image, $\mathbf{x}^{t}$, and of the reconstructed $N$ feature vectors, $\mathbf{\hat{x}}^{t}_{i}$, generated by the $N$ autoencoders, and determines the minimum:
\begin{equation}
\label{eq:dis}
			\centering
				d_{i*}^t= \min_{i=1,..,N} \sum_{l=1}^{D}||{x^{t}}(l)-{\hat{x}^{t}_{i}(l)}||_{2}^{2},
\end{equation}
where ${x^{t}}(l)$ and ${\hat{x}^{t}_i(l)}$ are the $l$-th elements of feature vector of the uploaded image and of its $i$-th reconstruction, respectively. 

The best reconstruction (i.e.~the minimum dissimilarity score, $d_{i*}^t$) is expected for the class of the uploaded image belongs to. If the minimum dissimilarity does not correspond to the class of the user who has uploaded the image, then that image should be labeled either as privacy-sensitive image of another user (and therefore blocked) or as non-privacy-sensitive image for any of the contributing users (and therefore shared). 

% Second, the probability density function (PDF) for a given $u_i$ is estimated as:
% \begin{equation}
% p(d|\mu_i,\sigma_i) =\dfrac{1}{\sqrt{2\pi\sigma_i^2}}e^-{{\dfrac{(d(\cdot)-\mu_i)^2}{2\sigma_i^2}}}
% \end{equation}
% where $\mu_i$ and $\sigma_i$ are the mean and standard deviation of the assumed normally distributed set of $d_i$. These parameters are locally calculated during the training of one-class autoencoders by users. 

To decide whether an image should be blocked or shared, we propose a {Guard} module that the service provider uses to determine the {\em privacy interval} of each autoencoder using its $\mu_i$ and $\sigma_i$, which were estimated at the end of the training phase (see Sec.~2.1). The privacy interval of each user $u_i$ is $[\mu_i-\alpha\sigma_i,\mu_i+\alpha\sigma_i]$, which defines the values of $d(\mathbf{x}^{t}, \mathbf{\hat{x}}^{t}_{i})$ for which $\mathbf{I}^{t}$ should be considered a privacy-sensitive image of user $i$. The value of  $\alpha \in [1,2]$ defines the desired confidence value (see Sec.~3). If the minimum distance is outside this interval, $\mathbf{I}^{t}$ is not considered as privacy-sensitive  and hence shared. Otherwise, $\mathbf{I}^{t}$ is blocked by the Guard. 

Note that each user can define multiple classes of privacy-sensitive images by training additional one-class autoencoders. Moreover, the proposed filter is scalable as each user individually trains a one-class classifier and this process takes place separately from that of other users. The number of users can, therefore, be increased in the test phase: new users train a one-class classifier on their privacy-sensitive images and upload the resulting parameters to the cloud to join the filter of the service provider. We will quantify the impact of increasing the number of users in the next section.

%%%%%%%%%%%%%%%%%%%%%%%%%%%%%%
\section{Evaluation}
\label{sec:evaluation}

To validate the performance of the proposed filter, we measure its overall accuracy, the per-class accuracy, the acceptance rate, the robustness to attacks and its scalability. As proof of concept that covers different classes a user might want to protect, we use three datasets (see Table~\ref{tab:mnist}):  IMDB~\cite{parkhi2015deep}, CIFAR-10~\cite{krizhevsky2009learning} and  MNIST~\cite{lecun1998gradient}.  We consider real limitations in these experiments such as different sizes of users' training dataset and limited training data. We assume $N=10$ users, each represented by one class.

%%%%%%%%%%%%%%%%%%%%%%%%%%%%%%
We randomly choose 10  celebrities from the {\bf IMDB} dataset (5 different sets) as users/classes. We assume that the privacy-sensitive data are their faces and therefore we detect and crop the faces~\cite{wolf2011face} with size $(3 \times 224 \times 224)$. The feature extractor of each user (i.e.~same for all users) $u_i$ extracts a $D=2,048$ dimension feature vector. We consider a smaller and a larger autoencoder which differ only in the size of their representation layer (32 and 132, respectively). Hence the size of input, representation and output layers are $2,048$, $32$ (or $132$) and $2,048$, respectively. To mimic the protection of  classes of images other than people or faces, we use  {\bf CIFAR-10}, which  contains 50K training and 10k test images ($3 \times 32 \times 32$) of 10 classes: airplane, automobile, bird, cat, deer, dog, frog, horse, ship and truck. Users obtain a $D=256$ dimension feature vector~\cite{chollet2015keras}\footnote{Modified version of ResNet for data of the size ($3 \times 32 \times 32$) (e.g. CIFAR-10)} from their images and hence the size of input, representation and output layers are $256$, $32$ and $256$, respectively. Finally, we consider {\bf MNIST} includes 60K training and 10K test $(1 \times 28 \times 28)$ images of 10 different handwritten digits. For example, $u_0$ has 5,923 images of digit $0$ and $u_1$ has 6,742 images of digit $1$ for training their one-class autoencoder. Due to the simplicity of this dataset, no feature extractor is used and the raw pixel values are given as input to the autoencoder. The size of input, representation and output layers of each user's one-class autoencoder are $784$, $32$ and $784$, respectively. 

We first quantify the {\em overall accuracy} as the ratio between the number of correctly predicted labels for all classes and the total number of test data of all of the classes; and the {\em per-class accuracy} as the ratio between the number of correctly predicted labels of each class and the total number of that class. 

The overall accuracy of the global filter is 97$\%$ and $80\%$ on MNIST and CIFAR-10, respectively; whereas the overall accuracy for IMDB with the smaller and larger autoencoder is 92$\%$ and 97$\%$, respectively. The per-class accuracy is shown in Fig.~\ref{fig:performance-cifar}. In MNIST the accuracy of digit 0 is over $99\%$, that of digit 9 is approximately $95\%$, as this digit shares similarity with digits 3, 4 and 7.
The proposed filter performs better on MNIST and IMDB, which are more related to our application, than on CIFAR-10, which has a high inter-class variability.

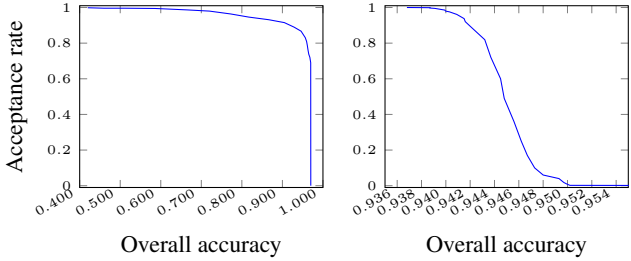
\begin{figure}[t!]
\subfloat{
\begin{tikzpicture}
\begin{axis}[
    tiny,
    width=0.56\linewidth,
    height=4cm,
    label style={font=\tiny},
    ylabel={Acceptance rate},
    ticklabel style = {font=\tiny},
    xlabel={Overall accuracy},
    x label style={font=\small},
    xticklabel style = {rotate=30,anchor=east},
    y label style={font=\small, at={(axis description cs:.2,.5)},anchor=south},
    ymajorgrids,
    yminorgrids,
    xmin=0.4, xmax=1,
    ymin=-0.01, ymax=1.01,
    x tick label style={
    /pgf/number format/.cd,
    fixed,
    fixed zerofill,
    precision=3
  },
   % x tick label style={precision =3},
    %ytick={0,20,40,60,80,100,120},
    legend pos=north west,
    ymajorgrids=false,
    grid style=dashed
]
\addplot[
    color=blue,
    ]
    coordinates {
(0.42,0.9986) (0.46,0.9960) (0.51,0.9960) (0.5856,0.9946) (0.6117,0.9920) (0.6648,0.9865) (0.72,0.9798) (0.7726,0.9637) (0.8170,0.9462) (0.8665,0.9314) (0.9047,0.9153) (0.9288,0.8898) (0.9461,0.8670) (0.9564,0.8306) (0.9597,0.8091) (0.9614,0.7876) (0.9633,0.7594) (0.9650,0.7380) (0.9677,0.7204) (0.9698,0.69) (0.97,0)
    };
\end{axis}
\end{tikzpicture}
}
\subfloat{
\begin{tikzpicture}
\begin{axis}[
    tiny,
    width=0.56\linewidth,
    height=4cm,
    label style={font=\tiny},
  %  ticklabel style = {font=\tiny},
    xlabel={Overall accuracy},
    xmin=0.935, xmax=0.955,
    ymin=-0.01, ymax=1.01,
    xticklabel style = {rotate=30,anchor=east},
    x label style={font=\small},
    x tick label style={
    /pgf/number format/.cd,
    fixed,
    fixed zerofill,
    precision=3,
  },
   % x tick label style={precision =3},
  %  ytick={0,0.2,0.4,0.6,0.8,1},
    legend pos=north west,
    ymajorgrids=false,
    grid style=dashed
]
\addplot[
    color=blue,
    ]
    coordinates {
   (0.9368, 1) (0.9387, 0.9990) (0.9387, 0.9980) (0.9387, 0.9970)     (0.9391,0.9950) (0.9399, 0.9857) (0.94, 0.9796) (0.9403, 0.9755) (0.9409,0.9612 ) (0.9415, 0.9377) (0.9416, 0.9204) (0.9420, 0.8949) (0.9432, 0.8193) (0.9437,0.7204 ) (0.9445, 0.6) (0.9448, 0.49) (0.9456,0.36 ) (0.9462, 0.25) (0.9467,0.17 ) (0.9473, 0.10) (0.9480, 0.06) (0.9493, 0.04) (0.9497,0.018 ) (0.9499,0.011 ) (0.9502, 0.002) (0.97,0)
    };
\end{axis}
\end{tikzpicture}
}
%\vspace{-10px}
 \caption{Accuracy vs.~acceptance rate on IMDB (left) and MNIST (right) when varying the confidence level $\alpha\in[1,2]$ in the Guard.}
%\vspace{-10px}
  \label{fig:block_no-block-mnist}
\end{figure}

Next, we quantify the {\em acceptance rate} of the Guard as ratio between the number of correctly shared non-privacy-sensitive images and the total number of uploaded non-privacy-sensitive images. We analyze the behavior of the global filter when a user, who contributed an autoencoder or is a new user, uploads non-privacy-sensitive images. We consider two scenarios: images that are substantially different from or are very similar to the privacy-sensitive images of registered users. The former scenario includes images of Fashion-MNIST~\cite{xiao2017fashion}, with ($1 \times 28 \times 28$) fashion products' images of 10 categories (t-shirt/top, trouser, pullover, dress, coat, sandal, shirt, sneaker, bag and ankle boot) in the MNIST experiment. The latter scenario includes faces with new identities (i.e.~not used in the training phase) of IMDB for the IMDB experiment. Fig.~\ref{fig:block_no-block-mnist} shows the relationship between sharing non-privacy-sensitive data (acceptance rate) and blocking privacy-sensitive images (overall accuracy). The larger the privacy interval in the Guard, the larger the overall accuracy and the lower the acceptance rate.  
%
%%%%%%%%%trade-off figure
%%%%%%%%%%%%%%%%%%%%%%%%%%%%%

\begin{figure}[t!]
\centering
\subfloat{\includegraphics[width=0.36\linewidth]{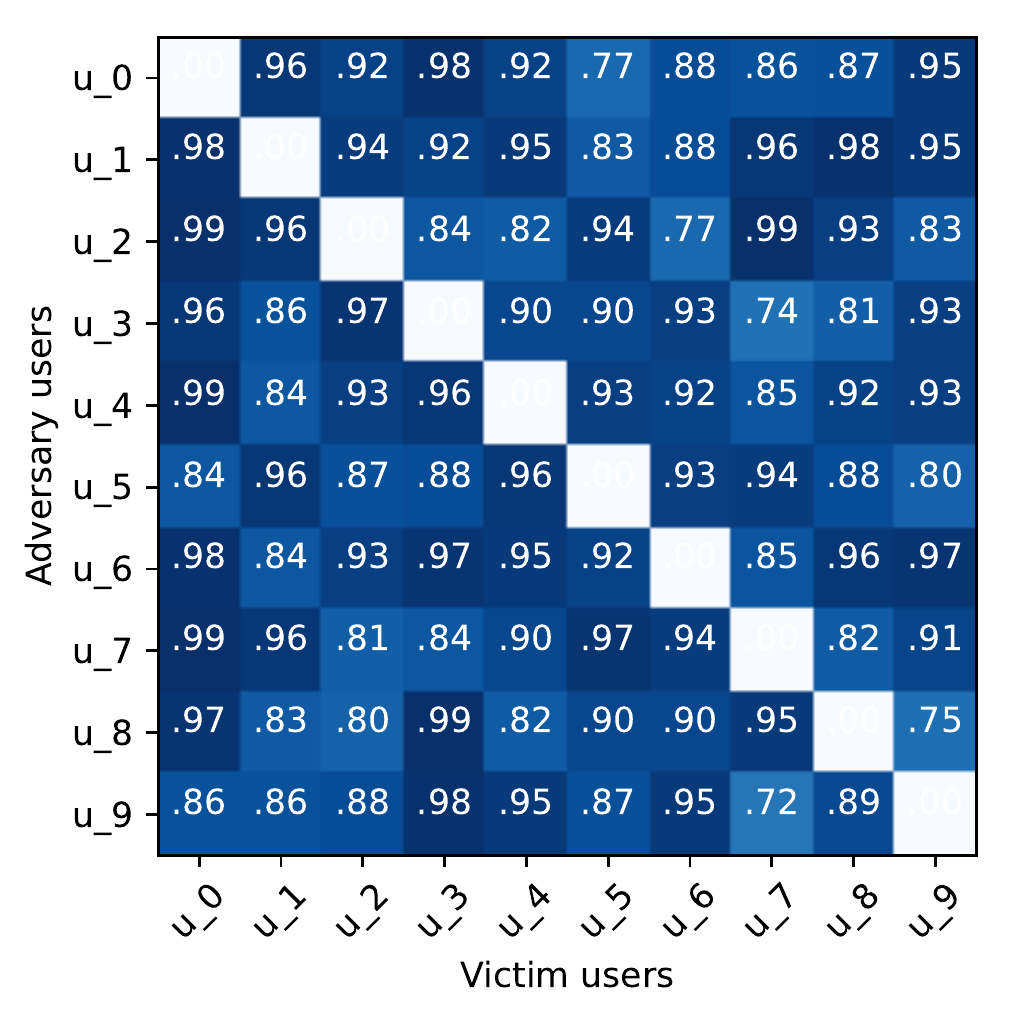}}\hspace*{-0.9em}
\subfloat{\includegraphics[width=0.32\linewidth]{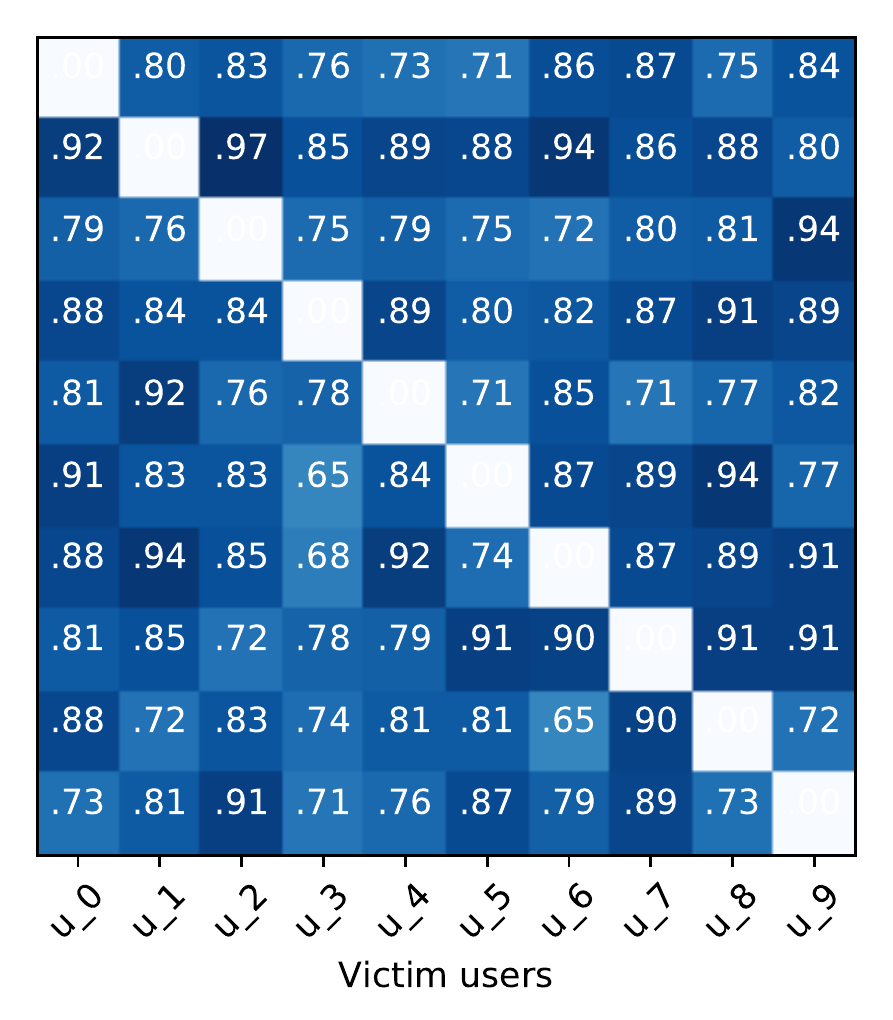}}\hspace*{-0.9em}
\subfloat{\includegraphics[width=0.32\linewidth]{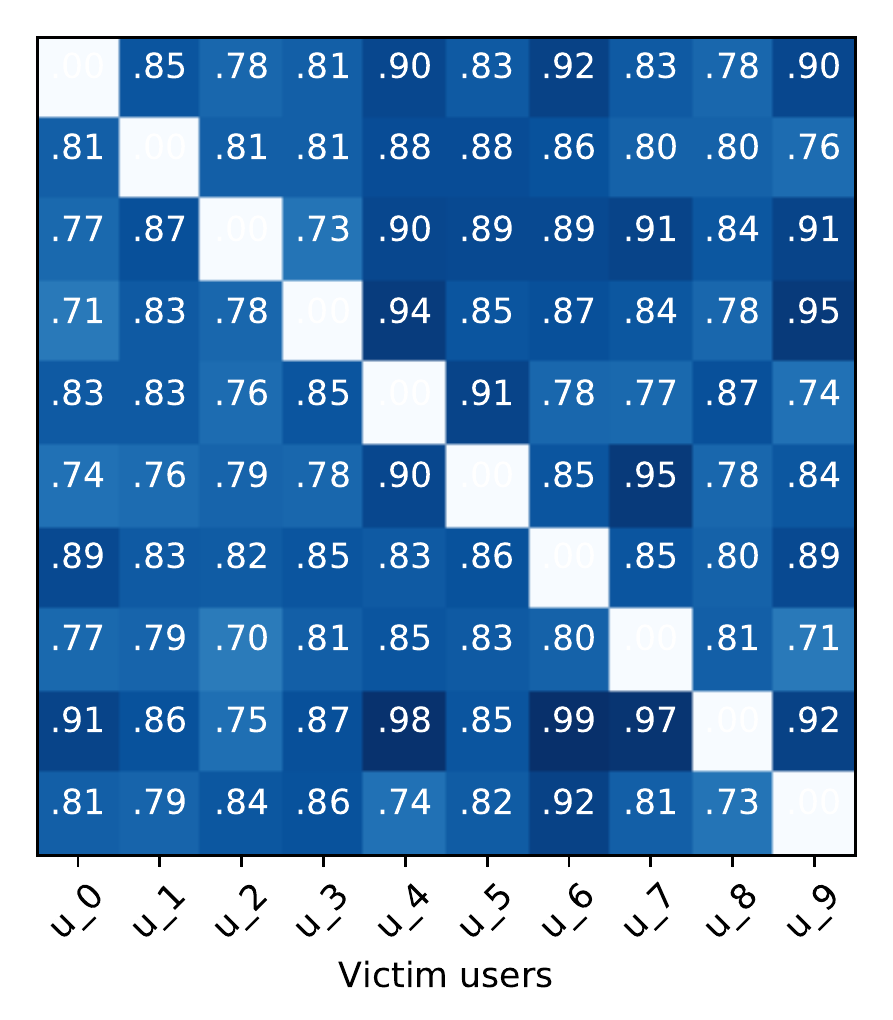}}\hspace*{-0.9em}
\caption{Accuracy in detecting an adversary user in IMDB (left), CIFAR-10 (middle) and MNIST (right) with the global filter.}
%\vspace{-10px}
\label{tab:adversary-mnist}
\end{figure}
Moreover, we analyze the {\em robustness} of the global filter against attacks when an adversary gains access to the data of another user and attempts to share their privacy-sensitive images publicly. We assume that the adversary user has full knowledge (i.e.~full access) of the training images of {\em some} users. This adversary user can train their local one-class autoencoder on the combination of his/her images with the available images of other users. Hence, the adversary user uploads the trained one-class classifier to the cloud in order to fool the service provider in sharing privacy-sensitive images of their targeted users.  To quantify robustness we consider each user as an adversary and compute the accuracy of detecting and blocking that user (Figure~\ref{tab:adversary-mnist}). The rows and columns represent adversary and victim users, respectively. For example, when we consider $u_3$ as an adversary and $u_2$ as victim user, it means that $u_3$ trains his/her local one-class autoencoder with images of $u_3$ and of $u_2$. Then $u_3$ uploads the trained one-class autoencoder to the cloud. Sharing privacy-sensitive images of each user by other users is blocked in IMDB dataset better than with the other two datasets. This result stems from the dissimilarity of data from different classes. For example in MNIST, digits 2 shares considerable similarities with digit 7, so $u_7$ succeeds in nearly 3 out of 10 attempts of sharing images of $u_2$ when  training a one-class autoencoder on data of digit 7 as well as digit 2.    

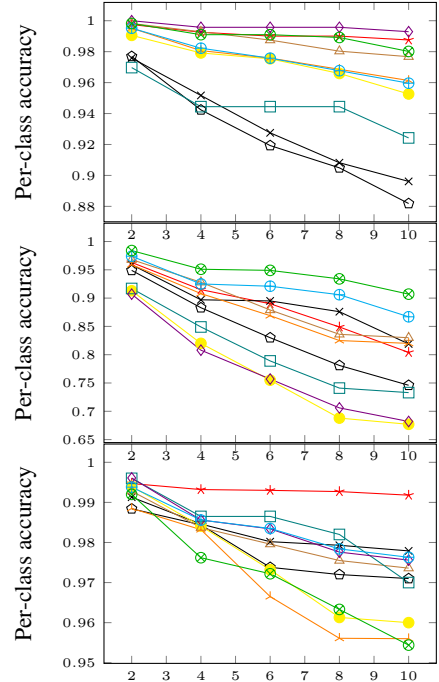
\begin{figure}[t]
\centering
\subfloat{
\begin{tikzpicture}
\begin{axis}[cycle list name=color list,
        tiny,
		width=6cm,
        height=4.5cm,
        ymax=1,
        enlarge y limits=0.1,legend style={at={(0.5,0.8)}},
         ylabel={Per-class accuracy},
        y label style={font=\small, at={(axis description cs:.1,.5)},anchor=south},]
\addplot+[mark=star]
    table[x=u,y=p0] {imdb.txt};\label{u0}
 %   \addlegendentry{$u_0$}
\addplot+[black,mark=x]
    table[x=u,y=p1] {imdb.txt};\label{u1}
%    \addlegendentry{$u_1$}
\addplot+[mark=pentagon]
    table[x=u,y=p2] {imdb.txt};\label{u2}
 %   \addlegendentry{$u_2$}
\addplot+[mark=*]
    table[x=u,y=p3] {imdb.txt};\label{u3}
%    \addlegendentry{$u_30$}
\addplot+[ mark=triangle]
    table[x=u,y=p4] {imdb.txt};\label{u4}
%    \addlegendentry{$u_4$}
\addplot+[ mark=square]
    table[x=u,y=p5] {imdb.txt};\label{u5}
%    \addlegendentry{$u_5$}
\addplot+[mark=Mercedes star]
    table[x=u,y=p6] {imdb.txt};\label{u6}
%    \addlegendentry{$u_6$}
\addplot+[mark=diamond]
    table[x=u,y=p7] {imdb.txt};\label{u7}
%    \addlegendentry{$u_7$}
\addplot+[mark=oplus]
    table[x=u,y=p8] {imdb.txt};\label{u8}
%    \addlegendentry{$u_8$}
\addplot+[mark=otimes]
    table[x=u,y=p9] {imdb.txt};\label{u9}
 %   \addlegendentry{$u_9$}
\end{axis}
\end{tikzpicture}
}\\
\vspace{-20px}
\subfloat{
\begin{tikzpicture}
\begin{axis}[cycle list name=color list,
        tiny,
		width=6cm,
        height=4.5cm,
        ymax=1,
        ylabel={Per-class accuracy},
        y label style={font=\small, at={(axis description cs:.1,.5)},anchor=south},
        enlarge y limits=0.1]
\addplot+[mark=star]
    table[x=u,y=p0] {cifar.txt};
\addplot+[black,mark=x]
    table[x=u,y=p1] {cifar.txt};
\addplot+[mark=pentagon]
    table[x=u,y=p2] {cifar.txt};
\addplot+[mark=*]
    table[x=u,y=p3] {cifar.txt};
\addplot+[ mark=triangle]
    table[x=u,y=p4] {cifar.txt};
\addplot+[ mark=square]
    table[x=u,y=p5] {cifar.txt};
\addplot+[mark=Mercedes star]
    table[x=u,y=p6] {cifar.txt};
\addplot+[mark=diamond]
    table[x=u,y=p7] {cifar.txt};
\addplot+[mark=oplus]
    table[x=u,y=p8] {cifar.txt};
\addplot+[mark=otimes]
    table[x=u,y=p9] {cifar.txt};
\end{axis}
\end{tikzpicture}
}\\
\vspace{-20px}
\subfloat{
\begin{tikzpicture}
\begin{axis}[cycle list name=color list,
        tiny,
		width=6cm,
        height=4.5cm,
        ymax=1,
        enlarge y limits=0.1,
        ylabel={Per-class accuracy},
        y label style={font=\small, at={(axis description cs:.1,.5)},anchor=south}]
\addplot+[mark=star]
    table[x=u0,y=p0] {mnist.txt};
\addplot+[black,mark=x]
    table[x=u1,y=p1] {mnist.txt};
\addplot+[mark=pentagon]
    table[x=u2,y=p2] {mnist.txt};
\addplot+[mark=*]
    table[x=u3,y=p3] {mnist.txt};
\addplot+[ mark=triangle]
    table[x=u4,y=p4] {mnist.txt};
\addplot+[ mark=square]
    table[x=u5,y=p5] {mnist.txt};
\addplot+[mark=Mercedes star]
    table[x=u6,y=p6] {mnist.txt};
\addplot+[mark=diamond]
    table[x=u7,y=p7] {mnist.txt};
\addplot+[mark=oplus]
    table[x=u8,y=p8] {mnist.txt};
\addplot+[mark=otimes]
    table[x=u9,y=p9] {mnist.txt};
\end{axis}
\end{tikzpicture}
}
%\vspace{-10px}
 \caption{The effect of increasing number of classes in IMDB (up), CIFAR-10 (middle) and MNIST (bottom) on the per-class accuracy of the global filter for $u_0$\ref{u0}, $u_1$\ref{u1}, $u_2$\ref{u2}, $u_3$\ref{u3}, $u_4$\ref{u4}, $u_5$\ref{u5}, $u_6$\ref{u6}, $u_7$\ref{u7}, $u_8$\ref{u8}, $u_9$\ref{u9} .}
  \label{fig:stab-mnist}
% \vspace{-10px}
\end{figure}

Finally, to evaluate the {\em scalability} of the framework, we analyze the per-class accuracy trend  when the number of classes increases. The per-class accuracy of the global filter for IMDB, CIFAR-10 and MNIST for a varying number of classes is shown in Figure~\ref{fig:stab-mnist}. The plots compare the accuracy of each class with different numbers of classes, i.e.~different combinations of {1, 3, 5, 7, 9} classes. The influence on performance when increasing number of classes on per-class accuracies in IMDB and MNIST is smaller than in CIFAR-10, because of its intra-class variability. The more similar the images in one class, the smaller the decrease in per-class and overall accuracy when the number of classes increases.

%%%%%%%%%%%%%%%%%%%%%%%%%%%%%Stability fig
\pgfplotscreateplotcyclelist{mycolorlist}{%
blue,every mark/.append style={fill=blue!80!black},mark=*\\%
red,every mark/.append style={fill=red!80!black},mark=square*\\%
brown!60!black,every mark/.append style={fill=brown!80!black},mark=otimes*\\%
black,mark=star\\%
blue,every mark/.append style={fill=blue!80!black},mark=diamond*\\%
red,densely dashed,every mark/.append style={solid,fill=red!80!black},mark=*\\%
brown!60!black,densely dashed,every mark/.append style={
solid,fill=brown!80!black},mark=square*\\%
black,densely dashed,every mark/.append style={solid,fill=gray},mark=otimes*\\%
blue,densely dashed,mark=star,every mark/.append style=solid\\%
red,densely dashed,every mark/.append style={solid,fill=red!80!black},mark=diamond*\\%
}

%%%%%%%%%%%%%%%%%%%%%%%%%%%%%%
\section{Conclusion}
\label{sec:conclusion}

We presented a filter that aims to prevent a user from sharing to a social networking website privacy-sensitive images of other users without their consent. The proposed filter enables a centralized classification without the need of sending the training data to the central server/cloud as the training phase is performed independently for each user on their specific class(es). 

This work is the first step in bringing one-class classifiers to distributed learning approaches in order to design a cloud-based filter with collaboration from the users with minimum computational costs and privacy loss. Giving sharing permission or blocking the uploading image is based on the comparison of the filter output and the user who has uploaded the image. 

Future work includes extending the validation of the proposed filter on larger datasets and on different types of privacy-sensitive data beyond images.

% Below is an example of how to insert images. Delete the ``\vspace'' line,
% uncomment the preceding line ``\centerline...'' and replace ``imageX.ps''
% with a suitable PostScript file name.
% -------------------------------------------------------------------------
% \begin{figure}[htb]

% \begin{minipage}[b]{1.0\linewidth}
%   \centering
%   \centerline{\includegraphics[width=8.5cm]{image1}}
% %  \vspace{2.0cm}
%   \centerline{(a) Result 1}\medskip
% \end{minipage}
% %
% \begin{minipage}[b]{.48\linewidth}
%   \centering
%   \centerline{\includegraphics[width=4.0cm]{image3}}
% %  \vspace{1.5cm}
%   \centerline{(b) Results 3}\medskip
% \end{minipage}
% \hfill
% \begin{minipage}[b]{0.48\linewidth}
%   \centering
%   \centerline{\includegraphics[width=4.0cm]{image4}}
% %  \vspace{1.5cm}
%   \centerline{(c) Result 4}\medskip
% \end{minipage}
% %
% \caption{Example of placing a figure with experimental results.}
% \label{fig:res}
% %
% \end{figure}

%\vfill
%\pagebreak


\end{document}